\documentclass{article}
\usepackage{ijcai23}
\usepackage{amsmath,graphicx}

\usepackage{times}
\usepackage{soul}
\usepackage{url}
\usepackage{amsfonts}
\usepackage{xcolor}
\usepackage{color, colortbl}
\definecolor{citecolor}{HTML}{2980b9}
\definecolor{linkcolor}{HTML}{c0392b}
\usepackage[hidelinks,breaklinks=true,bookmarks=false]{hyperref}
\usepackage[utf8]{inputenc}
\usepackage[small]{caption}
\usepackage{amsthm}
\usepackage{booktabs}
\usepackage{algorithm}
\usepackage{algorithmic}
\usepackage{adjustbox}
\usepackage{makecell}
\usepackage{multirow}
\usepackage{marvosym}
\makeatletter
  \newcommand\figcaption{\def\@captype{figure}\caption}
  \newcommand\tabcaption{\def\@captype{table}\caption}
\makeatother
\title{Joint-MAE: 2D-3D Joint Masked Autoencoders for\\ 3D Point Cloud Pre-training}

\author{
Ziyu Guo\textsuperscript{1}
\and
Renrui Zhang\textsuperscript{2}
\and
Longtian Qiu\textsuperscript{5}
\and
Xianzhi Li\textsuperscript{3}\and
Pheng-Ann Heng\textsuperscript{1, 4}\\
\affiliations
\textsuperscript{1} Department of Computer Science and Engineering, The Chinese University of Hong Kong\\
\textsuperscript{2} CUHK\ MMLab\\
\textsuperscript{3}Huazhong University of Science and Technology\\
\textsuperscript{4}Institute of Medical Intelligence and XR, The Chinese University of Hong Kong\\
\textsuperscript{5}ShanghaiTech University
\emails
\{zyguo, pheng\}@cse.cuhk.edu.hk,
xzli@hust.edu.cn
}

\begin{document}

\maketitle

\begin{abstract}
Masked Autoencoders (MAE) have shown promising performance in self-supervised learning for both 2D and 3D computer vision. However, existing MAE-style methods can only learn from the data of a single modality, i.e., either images or point clouds, which neglect the implicit semantic and geometric correlation between 2D and 3D. In this paper, we explore how the 2D modality can benefit 3D masked autoencoding, and propose \textbf{Joint-MAE}, a 2D-3D joint MAE framework for self-supervised 3D point cloud pre-training. Joint-MAE randomly masks an input 3D point cloud and its projected 2D images, and then reconstructs the masked information of the two modalities. For better cross-modal interaction, we construct our JointMAE by two hierarchical 2D-3D embedding modules, a joint encoder, and a joint decoder with modal-shared and model-specific decoders. On top of this, we further introduce two cross-modal strategies to boost the 3D representation learning, which are local-aligned attention mechanisms for 2D-3D semantic cues, and a cross-reconstruction loss for 2D-3D geometric constraints. By our pre-training paradigm, Joint-MAE achieves superior performance on multiple downstream tasks, e.g., 92.4\% accuracy for linear SVM on ModelNet40 and 86.07\% accuracy on the hardest split of ScanObjectNN.

\end{abstract}

\section{Introduction}
\vspace{0.1cm}
\label{sec:intro}
Self-supervised pre-training aims to learn more universal representations from unlabeled data that work across a wider variety of downstream tasks and datasets.
It has achieved great success in both 2D images and 3D point clouds with the main schemes including contrastive learning~\cite{afham2022crosspoint} and masked autoencoding~\cite{mae,gao2022convmae,pointmae,pointm2ae}. Recently, as the newly proposed paradigm, masked autoencoders have outperformed contrastive learning on multiple downstream tasks. It randomly masks a portion of input data and adopts a transformer encoder to extract the unmasked features. Then, a lightweight transformer decoder is utilized to reconstruct the information of the masked positions. 

\begin{figure}[t]
\vspace{15pt}
    \includegraphics[width=\linewidth]{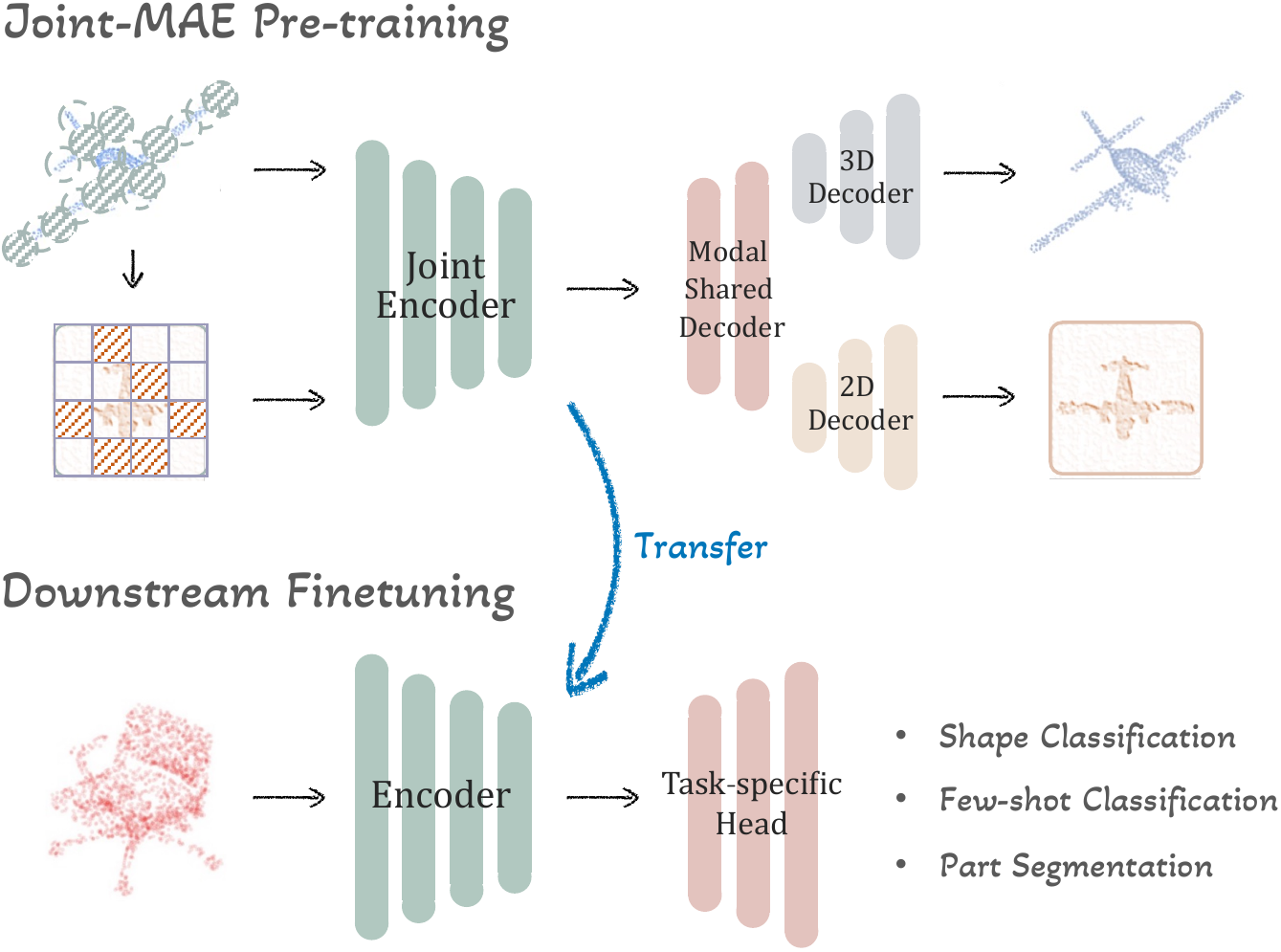}
  \caption{\textbf{The Pre-training and Fine-tuning Paradigms of Joint-MAE.} By joint masked autoencoding on both 3D and 2D, our Joint-MAE learns strong point cloud representations and exerts superior transferable capacity on 3D downstream tasks.}
  \label{teaser}
\end{figure}

In 2D modality, MAE~\cite{mae} and its follow-up work efficiently conduct 2D masked autoencoding with multi-scale convolution stages~\cite{gao2022convmae} and visible feature mimicking~\cite{gaomimic}. In 3D modality, Point-MAE~\cite{pointmae}, Point-M2AE~\cite{pointm2ae} and I2P-MAE~\cite{i2p} divide the point clouds into groups and directly reconstruct the masked coordinates in 3D space. At this point, existing methods have verified the effectiveness of MAE respectively in 2D and 3D. However, they can only process a single modality independently without utilizing their implicit correlations. Compared to the irregular and sparse point clouds, 2D images can provide dense and fine-grained visual signals concerning both geometries and semantics, which might assist the 3D representation learning during MAE pre-training. Therefore, we ask the question: \textit{Can we design a 2D-3D joint MAE framework to boost 3D point cloud pre-training by leveraging the information from 2D images?}

To address this issue, we propose \textbf{Joint-MAE}, a unified masked autoencoding framework that leverages cross-modal 2D knowledge to boost 3D representation learning. 
As shown in Figure~\ref{teaser}, by joint pre-training on both 2D images and 3D point clouds, our Joint-MAE can produce high-quality point cloud representation and present strong transferable capacity on 3D downstream tasks.
Specifically, to obtain the multi-modal input without using extra 2D data, we efficiently project the 3D point cloud into a 2D depth map from a random view. On top of the 2D-3D input data, we introduce two hierarchical modules to respectively embed 2D and 3D tokens and randomly mask them with a large ratio. Then, we concatenate the visible tokens of two modalities and feed them into a joint encoder, which interacts 2D-3D semantics via stacked transformer blocks. After that, we utilize a joint decoder to simultaneously reconstruct the masked point cloud coordinates and the image pixels, which consists of a modal-shared decoder and two modal-specific decoders.

Upon this unified paradigm, we further benefit 3D representation learning via two cross-modal strategies. First, in every transformer block of the joint encoder, we design a local-aligned attention mechanism for better interacting 2D-3D features. Only geometrically correlated 2D-3D tokens are valid for attention calculation, which can boost the point cloud representation learning with 2D fine-grained semantics. Second, besides the independent reconstruction losses of 2D and 3D, we propose an extra cross-reconstruction loss that provides additional 2D-3D geometric constraints.

The contributions of our paper are as follows:

\begin{itemize}
    \item We propose Joint-MAE, a 2D-3D joint MAE framework for self-supervised 3D point cloud pre-training, including two hierarchical 2D-3D embedding modules, a joint encoder, and a joint decoder with modal-shared and model-specific components.
    
    \item To better interact 2D-3D semantics and geometries, we introduce local-aligned attention mechanisms in the joint encoder and a cross-reconstruction loss for self-supervision.
    
    \item We evaluate Joint-MAE on various 3D downstream tasks, e.g., shape classification and part segmentation, where our approach shows leading performance compared with existing methods.
\end{itemize}
\section{Related Work}
\label{sec:related work}

\begin{figure*}[t!]
  \centering
    \includegraphics[width=\textwidth]{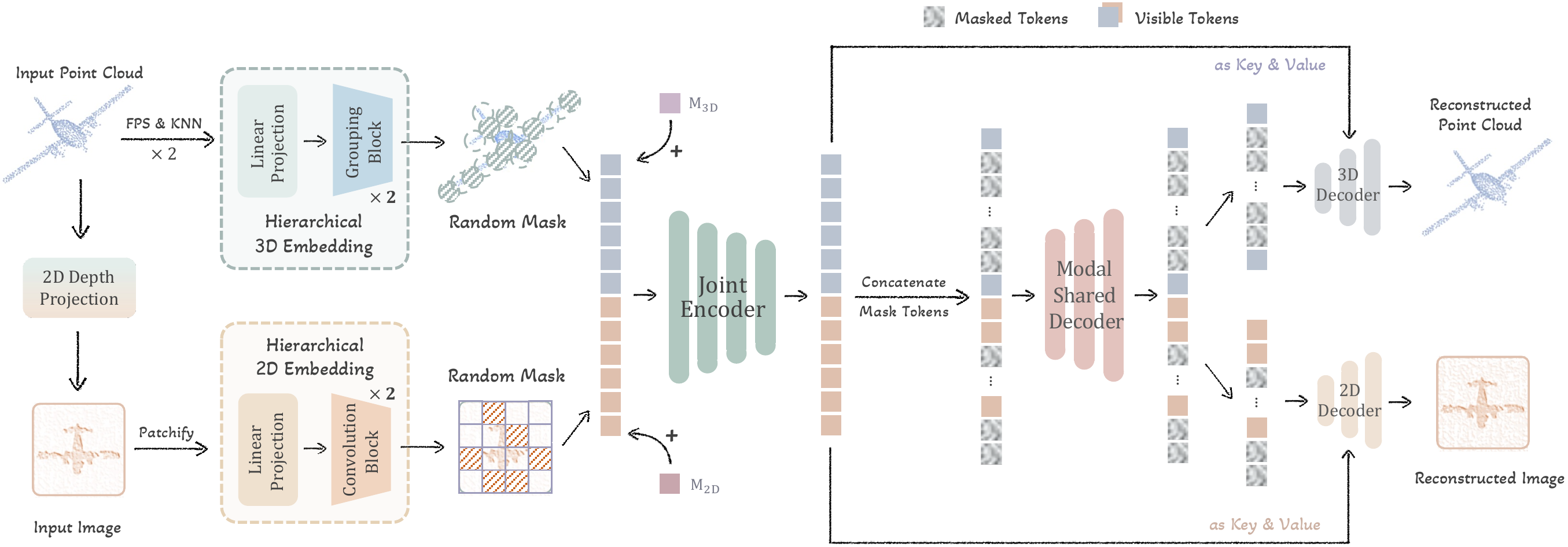}
   \caption{\textbf{The Pipeline of Joint-MAE,} which takes as input the 3D point cloud with its projected 2D image to conduct cross-modal masked autoencoding. To boost 3D representation learning by 2D-3D interaction, Joint-MAE consists of two hierarchical 2D-3D embedding modules, a joint encoder, and a joint decoder with modal-shared and modal-specific components.}
    \label{pipeline}
\end{figure*}

\paragraph{Pre-training in 3D Vision.}
Recently, supervised learning in 3D vision~\cite{qi2017pointnet,qi2017pointnet++,guo2021pct,dgcnn,zhang2023nearest} has achieved promising performance with delicate network designs. However, they are confined to limited data domains~\cite{modelnet40,scanobjectnn} that they are trained on, without satisfactory generalization ability.
Therefore, self-supervised learning emerges in this context, and has significantly enhanced the 3D transfer learning. That is, big models with abundant pre-trained knowledge show promising results after fine-tuning.
Meanwhile, the development of generative models for 3D point cloud~\cite{xie2016theory,xie2018learning,xie2020generative,xie2021generative,nichol2022point} promotes exploration of the pre-training paradigm on large-scale data.
Most self-supervised pre-training methods~\cite{poursaeed2020self} include an encoder and a decoder to firstly transform the input point clouds into latent representations, and then recover them into the original data form. Other works~\cite{rao2020global} conducts self-supervised pre-training via contrastive learning. Also, inspired by masked image modeling~\cite{mae}, a series of works for masked point modeling~\cite{maskpoint,pointbert} are put forward. Therein, Point-MAE~\cite{pointmae} conduct masked autoencoding for 3D point clouds directly. Point-M2AE~\cite{pointm2ae} and I2P-MAE~\cite{i2p} further adopt hierarchical MAE transformers. Our Joint-MAE is also based on the masked autoencoding framework, and further utilizes the correlated information between 2D and 3D modalities to build a more powerful 3D self-supervised learner.

\paragraph{Multi-Modality Learning.}
Multi-modality learning, namely, cross-modal learning, aims at learning from signals of multiple modalities at the same time, which achieves more robust performance than the single-modality learning method.
Most works~\cite{desai2021virtex} have proved the powerful capability of multi-modality pre-training. CLIP~\cite{clip} bridges the gap between textual space and image space through contrastive learning. Based on CLIP, some works~\cite{gao2021clip,tipadapter,guo2022calip} further propose more advanced structures to promote the performance, and adopt CLIP's pre-trained knowledge on different downstream tasks, e.g., PointCLIP V1~\cite{zhang2021pointclip}, V2~\cite{pointclipv2} for 3D learning, and DepthCLIP~\cite{depthclip} for depth estimation. ~\cite{zhang2021self} conducts a point-voxel joint pre-training design, and CrossPoint~\cite{afham2022crosspoint} proposes an image-point contrastive learning network. Also, via filter inflation, ~\cite{xu2021image2point} utilizes pre-trained 2D knowledge on 3D point cloud understanding. 
Inspired by the success of MAE pre-training independently in 2D and 3D vision~\cite{mae,gao2022convmae,pointmae,pointm2ae}, we expect to fully incorporate MAE pre-training and multi-modality learning to unleash their potientials for 3D representation learning.

\section{Method}
\label{sec:method}
The overall pre-training pipeline of Joint-MAE is shown in Figure~\ref{pipeline}. 
Given an input 3D point cloud, we first generate 2D inputs via efficient depth map projection, and adopt 2D-3D embedding modules for initial tokenization (Section~\ref{embed}). Then, we feed the visible tokens into a 2D-3D transformer, which consists of a joint encoder and a joint decoder (Section~\ref{transformer}). Finally, we further introduce two cross-modal strategies, i.e., the local-aligned attention mechanism (Section~\ref{attention}) and the cross-reconstruction loss (Section~\ref{loss}).




\subsection{2D-3D Data Embedding}
\label{embed}

\paragraph{2D Depth Projection from 3D.}
\label{project}
To conduct efficient multi-modal learning for 3D point cloud pre-training, we require no additional real-world 2D images and simply project 3D point clouds into depth maps as the 2D input.
For an input point cloud $P \in R^{N\times3}$ with $N$ points, we select a random view at every iteration for augmentation and perspectively project~\cite{simpleview,zhang2021pointclip} 3D points onto a pre-defined image plane as $I$, formulated as
\begin{align}
\label{project}
\begin{split}
    & I = \operatorname{Project}(P),\ \ \text{where}\ I \in R^{H \times W \times 1}.\\
\end{split}
\end{align}
In this way, we not only discard the time-consuming rendering, but also preserve the 2D-3D geometric correlations.


\paragraph{Hierarchical 2D-3D Tokenization.}
Unlike MAE~\cite{mae} and Point-MAE~\cite{pointmae} that tokenize the 2D and 3D data with a constant resolution, we propose hierarchical 2D-3D embedding modules to extract multi-scale features independently for the two modalities. For the 3D branch, we first adopt two cascaded Farthest Point Sampling (FPS) with K Nearest Neighbors (k-NN) to acquire the multi-scale representations of point cloud, which contain $N/8$ and $N/32$ points respectively. After transforming the raw points via a linear projection layer, we utilize two grouping blocks to hierarchically tokenize the point cloud according to the obtained multi-scale representations, which produces the $C$-dimensional 3D tokens, $T_{3D}\in R^{N/32 \times C}$. The grouping block consists of a mini-PointNet~\cite{qi2017pointnet} for aggregating local features within each k neighbors. For the 2D branch, we first patchify the depth map with a patch size $4\times4$ and utilize a linear projection layer for high-dimensional embedding. Then, two $3\times3$ convolution blocks are adopted to progressively tokenize the 2D image as the $C$-dimensional 2D tokens with a downsample ratio $1/16$, denoted as $T_{2D}\in R^{HW/256 \times C}$. We formulate the 2D-3D embedding modules as
\begin{align}
\begin{split}
    T_{3D} = \operatorname{Embed}_{3D}(P),\ \ T_{2D} = \operatorname{Embed}_{2D}(I).\\
\end{split}
\end{align}

\paragraph{2D-3D Masking.}
On top of this, we randomly mask the 2D and 3D tokens respectively with a large ratio, e.g., 75\%. This builds a challenging pre-text task for self-supervised learning by masked auto-encoding. We denote the visible tokens as $T^v_{2D}$ and $T^v_{3D}$, which are fed into a proposed 2D-3D joint transformer for multi-modal reconstruction.



\subsection{2D-3D Joint Transformer}
\label{transformer}
The transformer consists of a joint encoder and a joint decoder. The former encodes cross-modal features upon the visible 2D-3D tokens. The latter reconstructs the masked 2D-3D information with a modal-shared decoder and two modal-specific decoders.

\subsubsection{Joint Encoder}
We concatenate the visible tokens of two modalities along the token dimension as the input for joint encoder, which is composed of 12 transformer blocks with self-attention layers for 2D-3D feature interaction. To distinguish the modality characters during attention mechanisms, we introduce the modality encodings by using two learnable tokens and randomly initializing them before pre-training. The two modality tokens, $M_{2D}, M_{3D}\in  \mathbb{R}^{1\times C}$, are respectively added to all the 2D and 3D tokens, which can provide modal-specific cues for attention layers. We formulate it as
\begin{align}
\begin{split}
    E^v_{2D\text{-}3D} = \operatorname{JointEnc}\big(\operatorname{Concat}(T^v_{3D} + M_{3D}, T^v_{2D} + M_{2D})\big),\nonumber\\
\end{split}
\end{align}
where $E^v_{2D\text{-}3D}$ denotes the visible 2D-3D features of the input point cloud. By the multiple attention layers, the 3D tokens can adaptively aggregate informative 2D semantics for 2D-3D interaction, and obtain a cross-modal understanding of the point cloud.


\subsubsection{Joint Decoder}
After the encoding of visible 2D-3D tokens, we propose a joint decoder for multi-modal reconstruction of the masked information, which consists of a modal-shared decoder and two modal-specific decoders.

\begin{figure}[t]
    \includegraphics[width=\linewidth]{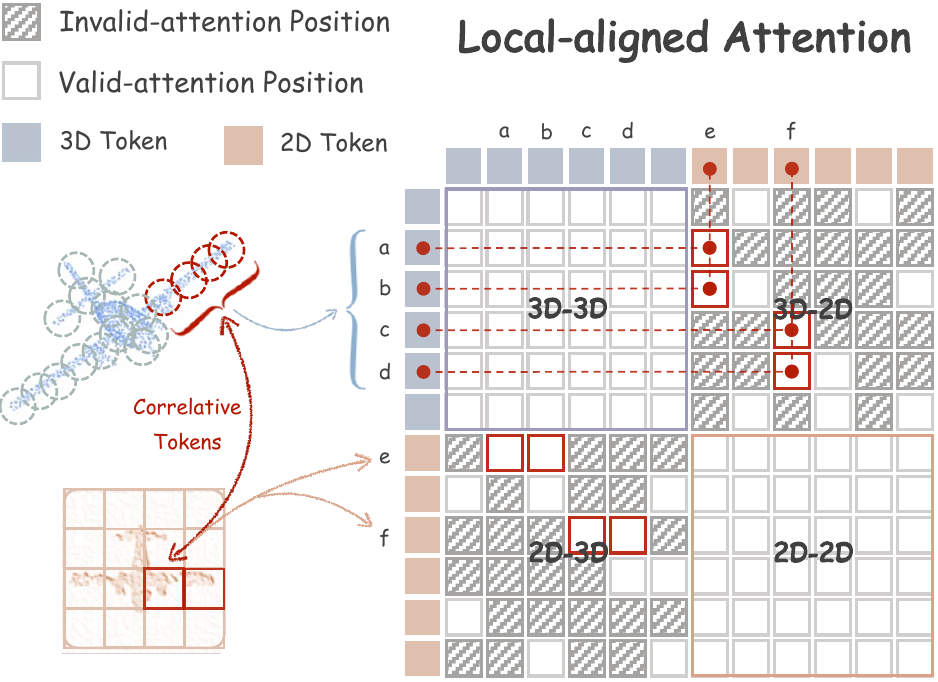}
  \caption{\textbf{Local-aligned Attention.} For self-attention mechanisms of the joint encoder, we only interact cross-modal features between the geometrically correlative 2D-3D tokens (highlighted in red). This contributes to more fine-grained and intensive feature learning.}
  \label{attn}
\end{figure}

\paragraph{Modal-shared Decoder.}
We adopt two learnable $C$-dimensional mask tokens in the decoder respectively for 2D and 3D, which are shared for all masked positions, denoted as $T^m_{3D}$ and $T^m_{2D}$. We concatenate them with the encoded 2D-3D tokens $E^v_{2D\text{-}3D}$ along the token dimension, and feed them into the model-shared decoder, which contains 2 transformer blocks with self-attention layers. We formulate it as 
\begin{align}
\label{share}
\begin{split}
    D_{2D\text{-}3D} = \operatorname{ShareDec\big(Concat}(E^v_{2D\text{-}3D}, T^m_{3D}, T^m_{2D})\big),\\
\end{split}
\end{align}
where $D_{2D\text{-}3D}$ denotes the decoded 2D-3D features. By the model-shared decoder, the 3D mask tokens $T^m_{3D}$ can simultaneously capture information from both 2D and 3D visible tokens. In this way, the reconstruction of masked 3D point clouds can be benefited from both the visible 3D geometries and the additional 2D patterns, which contributes to better 3D representation learning and vice versa.

\paragraph{Modal-specific Decoders.} 
After the modal-shared decoder, we divide $D_{2D\text{-}3D}$ along the token dimension as the 2D and 3D tokens, denoted as $D_{2D}$ and $D_{3D}$.
Then, they are fed into two concurrent decoders for specific 2D and 3D decoding. This enables the network to respectively focus on the unique characteristics of 2D and 3D for their reconstruction.
Importantly, we adopt cross-attention layer in the 1-block specific decoders, where $D_{2D}, D_{3D}$ serve as queries and $E^v_{2D}, E^v_{3D}$ serve as keys/values. $E^v_{2D}, E^v_{3D}$ are the token-wise division of $E^v_{2D\text{-}3D}$ respectively for 2D and 3D modalities, which can complement more model-specific cues for the mask tokens during the cross-attention mechanism. We formulate them as
\begin{align}
\label{de}
\begin{split}
    D'_{3D} &= \operatorname{SpeDec_{3D}}(D_{3D}, E^v_{3D})\\
    D'_{2D} &= \operatorname{SpeDec_{2D}}(D_{2D}, E^v_{2D})
\end{split}
\end{align}
where $D'_{2D}, D'_{3D}$ represent the modal-specific decoded features for 2D and 3D.

\subsection{Local-aligned Attention}
\label{attention}
To further enhance the 2D-3D feature interaction, we propose a local-aligned attention mechanism in every transformer block of the joint encoder. Figure~\ref{attn} illustrates its methodology. During the attention layer, all tokens are mutually visible within their own modalities: every 3D token conducts attention calculation with each other, and 2D likewise. However, only the geometrically correlated 2D and 3D tokens would be calculated the attention scores for feature interaction (Valid-attention Position), and others would not (Invalid-attention Position). We regard a 3D token is geometrically correlated with a 2D token if its 3D center point is projected onto the pixels within the 2D token on the image plane. As shown in the figure, the 3D tokens `a, b, c, d' are correlated with the 2D tokens `e, f, g', both of which represent the same part-wise features of the object, i.e., the wing of an airplane. We only enable the attention scores between these correlated 2D-3D tokens to be valid. By such local-aligned attention, the cross-modal interaction is effectively constrained into different local regions, which guides the tokens to intensively aggregate fine-grained features from the other modality and avoids the long-range distractions of non-relevant regions.



\begin{figure}[t]
\vspace{0.1cm}
\centering
    \includegraphics[width=0.95\linewidth]{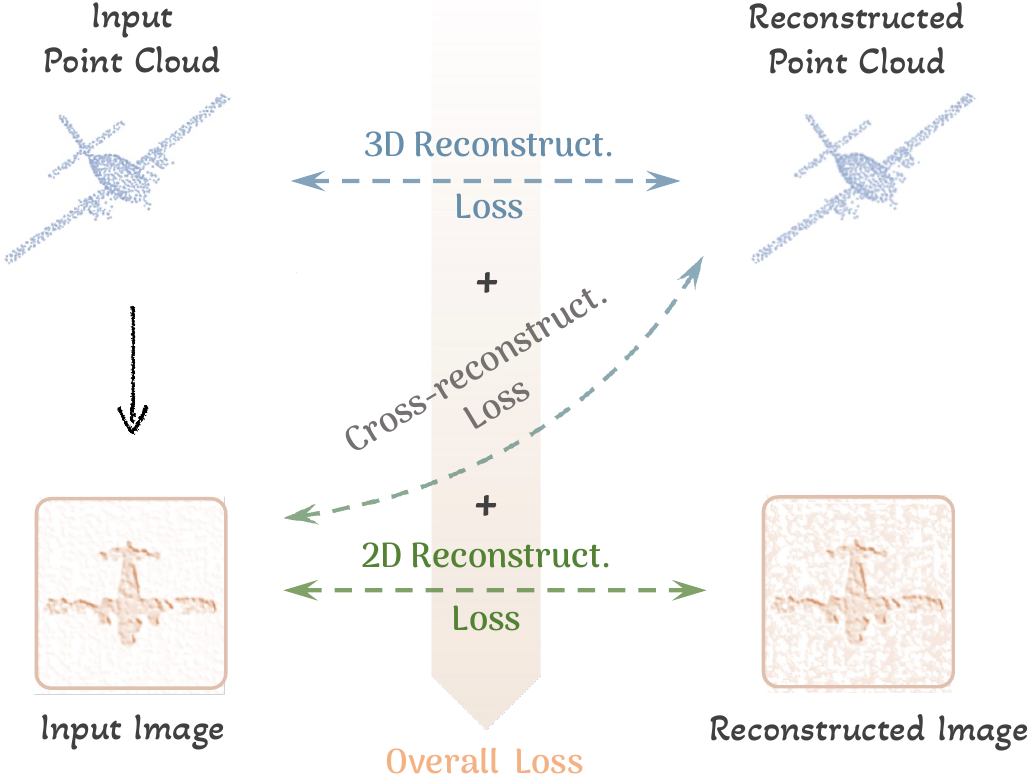}
  \caption{\textbf{2D-3D Reconstruction Loss.} We apply three losses for Joint-MAE pre-training, which are 2D and 3D reconstruction losses, along with a cross-reconstruction loss that provides additional 2D-3D geometric constrains.}
  \label{loss_fig}
\end{figure}

\subsection{2D-3D Reconstruction Loss}
\label{loss}
As shown in Figure~\ref{loss_fig}, we adopt two reconstruct heads of simple linear layers to respectively recover the masked 2D pixels and 3D coordinates following MAE~\cite{mae} and Point-MAE~\cite{pointmae}. We adopt Mean Squared Error (MSE) and $l_2$ Chamfer Distance (CD)~\cite{fan2017point} as two loss functions of 2D and 3D reconstruction, i.e.,
\begin{align}
\begin{split}
    L_{3D} &= \operatorname{CD}\big(\operatorname{RecHead_{3D}}(D'_{3D}),\ P\big),\\
    L_{2D} &= \operatorname{MSE}\big(\operatorname{RecHead_{2D}}(D'_{2D}),\ I\big).
\end{split}
\end{align}
Besides, for better 2D-3D geometric alignment, we propose an extra cross-reconstruction loss between the reconstructed point cloud and the 2D input $I$, the initially projected depth map. Specifically, we project the reconstructed point cloud by the same view of $I$ and then calculate the MSE loss as
\begin{align}
\label{crossloss}
\begin{split}
    L_{3D\text{-}2D} = \operatorname{MSE}\Big(\operatorname{Project\big(\operatorname{RecHead_{3D}}}(D'_{3D})\big),\ I\Big).
\end{split}
\end{align}
Such cross-reconstruction loss can better self-supervise the spatial structure of reconstructed point clouds and enforce the 3D-2D geometric alignment in the 2D space.
Then, the overall loss for Joint-MAE pre-training is formulated as
\begin{align}
\begin{split}
    L_{overall} = L_{2D} + L_{3D} + L_{3D\text{-}2D}.
\end{split}
\end{align}
The three self-supervisory signals can well interact 2D-3D features from both semantics and geometries, which effectively boosts the representation learning for 3D poitn clouds.


\begin{table}[t!]
\centering
\small
\begin{tabular}{lc}
\toprule
	Method &Accuracy (\%)\\
    \cmidrule(lr){1-1} \cmidrule(lr){2-2}
    SO-Net~\cite{sonet}   &87.3  \\
    FoldingNet~\cite{foldingnet}  &88.4 \\
	VIP-GAN~\cite{vipgan}   &90.2 \\
	\cmidrule(lr){1-2}
	DGCNN + Jiasaw~\cite{jiasaw}  &90.6  \\
	DGCNN + OcCo~\cite{occo}  &90.7  \\
    DGCNN + CrossPoint~\cite{afham2022crosspoint}  &91.2 \\
    \cmidrule(lr){1-2}
    Transformer + OcCo~\cite{occo} &89.6 \\
    Point-BERT~\cite{pointbert}  &87.4\\
    Point-MAE~\cite{pointmae}  &91.0\\
	\rowcolor{blue!5}\textbf{Joint-MAE}  &\textbf{92.4}\vspace{0.1cm}\\
\bottomrule
\end{tabular}
\tabcaption{\textbf{Linear SVM Performance.} We report shape classification results of different self-supervised pre-training methods on ModelNet40 \protect\cite{modelnet40} via linear SVM.}
\label{svm}
\end{table}

\begin{table}[t!]
\vspace{0.2cm}
\centering
\small
	\begin{tabular}{lc}
	\toprule
		\makecell*[c]{Method} &Accuracy (\%)\\
		\cmidrule(lr){1-1} \cmidrule(lr){2-2} 
         PointNet~\cite{qi2017pointnet} &89.2 \\ 
        PointNet++~\cite{qi2017pointnet++} &90.5 \\
        DGCNN~\cite{dgcnn} &92.9 \\
        PCT~\cite{guo2021pct} &93.2 \\
        Point Transformer~\cite{pointtransformer} &93.7 \\
        \cmidrule(lr){1-1} \cmidrule(lr){2-2}
        {Transformer + OcCo}~\cite{occo} &92.1\\
        {Point-BERT}~\cite{pointbert} &93.2\\
        {Point-MAE}~\cite{pointmae} &93.8\\
        {MaskPoint}~\cite{maskpoint} &93.8\vspace{0.1cm}\\
        \rowcolor{blue!5}\textbf{Joint-MAE} &\textbf{94.0}\vspace{0.1cm}\\
	\bottomrule
	\end{tabular}
\tabcaption{\textbf{Shape Classification on ModelNet40 
\protect\cite{modelnet40}.} We report the shape classification accuracy (\%) of Joint-MAE finetuned on ModelNet40 dataset.}
\label{modelnet_cls}
\end{table}

\section{Experiments}
\label{sec:experiments}

In this section, we evaluate the performance of Joint-MAE on various downstream tasks, i.e., shape classification, few-shot classification, and part segmentation.

\subsection{Pre-training.}
\label{pretrain}
\paragraph{Settings.}
Following existing works~\cite{pointmae,pointm2ae}, we pre-train our Joint-MAE on ShapeNet~\cite{chang2015shapenet}, which covers 57,448 3D shapes of 55 categories. 
The input point number $N$ is set as 2,048 and the depth map size $H\times W$ is set as 224$\times$224. We adopt a feature dimension $C$ as 384. Please refer to the Supplementary Material for detailed implementation details. \textbf{\textit{After Joint-MAE pre-training, we utilize the pre-trained encoder along with the hierarchical 3D embedding module for 3D downstream tasks, where we only preserve the 3D branch, i.e., only 3D point clouds as input without projecting into depth maps.}}

\begin{table}[t]
\small
\centering
	\begin{tabular}{lccc}
	\toprule
		\makecell*[c]{Method} &OBJ-BG &OBJ-ONLY &PB-T50-RS\\
		\cmidrule(lr){1-1} \cmidrule(lr){2-2} \cmidrule(lr){3-3} \cmidrule(lr){4-4}
	    PointNet
     &73.3 &79.2 &68.0\\
	    PointNet++
     &82.3 &84.3 &77.9\\
	    DGCNN
     &82.8 &86.2 &78.1\\
	    PointCNN
     &86.1 &85.5 &78.5\\
	    GBNet
     &- &-  &80.5\\
	    SimpleView
     &- &-  &80.5\\
	    PRANet
     &- &- &81.0\\
	    MVTN
     &- &-  &82.8\\
	    PointMLP
     &- &-  &85.2\\
	   \cmidrule(lr){1-4}
	    Transformer
     &79.86 &80.55 &77.24  \\
	    Transformer + OcCo
     &84.85 &85.54 &78.79\\
	    Point-BERT
     &87.43 &88.12 &83.07\\
            MaskPoint &89.30 &88.10 &84.30 \\
	    Point-MAE
     &90.02 &88.29 &85.18\vspace{0.1cm}\\
	    \rowcolor{blue!5} \textbf{Joint-MAE} &\textbf{90.94} &\textbf{88.86}  &\textbf{86.07}\vspace{0.1cm}\\
	     \textit{Improvement} &\textcolor{violet}{+0.92} &\textcolor{violet}{+0.57} &\textcolor{violet}{+0.89} \\
	  \bottomrule
	\end{tabular}
\tabcaption{\textbf{Shape Classification on ScanObjectNN \protect\cite{scanobjectnn}}. We report the shape classification accuracy (\%) on the three splits of ScanObjectNN.}
\label{scan_cls}
\end{table}

\paragraph{Performance of Linear SVM.}
We first evaluate the Linear SVM performance of Joint-MAE on the well-known ModelNet40 dataset~\cite{wu20153d}. Following CrossPoint~\cite{afham2022crosspoint}, we utilize the pre-trained encoder and freezes its weights to extract the 3D point cloud features. We report the shape classification accuracy in Table~\ref{svm}. 
 As shown, our Joint-MAE outperforms all existing self-supervised pre-training methods and surpasses the second-top Point-MAE~\cite{pointmae} with \textbf{+1.4\%} improvement, indicating the superior 3D representation assested by the 2D modality. 

\begin{figure*}[t!]
\centering
\begin{minipage}[t!]{0.75\linewidth}
\small
\centering
	\begin{tabular}{lc c c c c}
	\toprule
		\makecell*[c]{\multirow{2}*{Method}} &\multicolumn{2}{c}{5-way} &\multicolumn{2}{c}{10-way}\\
		 \cmidrule(lr){2-3} \cmidrule(lr){4-5}
		 &10-shot &20-shot &10-shot &20-shot\\
		 \cmidrule(lr){1-1} \cmidrule(lr){2-5}
		PointNet~\cite{qi2017pointnet} &52.0\ $\pm$\ 3.8 &57.8\ $\pm$\ 4.9 &46.6\ $\pm$\ 4.3 &35.2\ $\pm$\ 4.8 \\
		PointNet + OcCo~\cite{occo} &89.7\ $\pm$\ 1.9 &92.4\ $\pm$\ 1.6 &83.9\ $\pm$\ 1.8 &89.7\ $\pm$\ 1.5 \\
		PointNet + CrossPoint~\cite{afham2022crosspoint} &90.9\ $\pm$\ 4.8 &93.5\ $\pm$\ 4.4 &84.6\ $\pm$\ 4.7 &90.2\ $\pm$\ 2.2 \\
		\cmidrule(lr){1-5}
		DGCNN~\cite{dgcnn} &31.6\ $\pm$\ 2.8 &40.8\ $\pm$\ 4.6 &19.9\ $\pm$\ 2.1 &16.9\ $\pm$\ 1.5 \\
		DGCNN + OcCo~\cite{occo} &90.6\ $\pm$\ 2.8 &92.5\ $\pm$\ 1.9 &82.9\ $\pm$\ 1.3 &86.5\ $\pm$\ 2.2 \\ 
		DGCNN + CrossPoint~\cite{afham2022crosspoint} &92.5\ $\pm$\ 3.0 &94.9\ $\pm$\ 2.1 &83.6\ $\pm$\ 5.3 &87.9\ $\pm$\ 4.2 \\
		\cmidrule(lr){1-5}
	    Transformer~\cite{transformer} &87.8\ $\pm$\ 5.2 &93.3\ $\pm$\ 4.3 &84.6\ $\pm$\ 5.5 &89.4\ $\pm$\ 6.3\\
		Transformer + OcCo~\cite{occo} &94.0\ $\pm$\ 3.6 &95.9\ $\pm$\ 2.3 &89.4\ $\pm$\ 5.1 &92.4\ $\pm$\ 4.6\\
		Point-BERT~\cite{pointbert} &94.6\ $\pm$\ 3.1 &96.3\ $\pm$\ 2.7 &91.0\ $\pm$\ 5.4 &92.7\ $\pm$\ 5.1\\
            MaskPoint~\cite{maskpoint}  &95.0\ $\pm$\ 3.7 &97.2\ $\pm$\ 1.7 &91.4\ $\pm$\ 4.0 &93.4\ $\pm$\ 3.5\\
	    Point-MAE~\cite{pointmae} &96.3\ $\pm$\ 2.5 &97.8\ $\pm$\ 1.8 &92.6\ $\pm$\ 4.1 &95.0\ $\pm$\ 3.0\\
		\rowcolor{blue!5} \textbf{Joint-MAE} &\textbf{96.7\ $\pm$\ 2.2}& \textbf{97.9\ $\pm$\ 1.8}&\textbf{92.6\ $\pm$\ 3.7} &\textbf{95.1\ $\pm$\ 2.6}\vspace{0.1cm}\\
	\bottomrule
	\end{tabular}
\end{minipage}
\tabcaption{\textbf{Few-shot Classification on ModelNet40\protect\cite{modelnet40}}. We report the average accuracy ($\%$) and standard deviation ($\%$) of 10 independent experiments.} 
\label{fewshot}
\vspace{-0.4cm}
\end{figure*}

\subsection{Downstream Tasks}
After pre-training, we fine-tune the 3D branch of Joint-MAE on multiple 3D downstream tasks, i.e., shape classification, few-shot classification, and part segmentation. In each task, we discard the 2D branch with the decoder, and equip the pre-trained encoder by the task-specific heads.

\begin{figure}[t!]
\vspace{0.2cm}
\small
\centering
\centering
	\begin{tabular}{lccc}
	\toprule
		Method &mIoU$_C$ &mIoU$_I$\\
		\cmidrule(lr){1-1} \cmidrule(lr){2-2} \cmidrule(lr){3-3} 
	    PointNet~\cite{qi2017pointnet} &80.39 &83.70 \\
	    PointNet++~\cite{qi2017pointnet++} &81.85 &85.10 \\
	    DGCNN~\cite{dgcnn} &82.33 &85.20 \\
	    \cmidrule(lr){1-3}
	    Transformer~\cite{transformer} &83.42 &85.10 \\
	    {Transformer + OcCo}~\cite{occo} &83.42 &85.10 \\
	    Point-BERT~\cite{pointbert} &84.11 &85.60 \\
            MaskPoint~\cite{maskpoint} &84.40 &86.00 \\
	    Point-MAE~\cite{pointmae} &- &86.10 \vspace{0.05cm}\\
	    \rowcolor{blue!5} \textbf{Joint-MAE} &\textbf{85.41} &\textbf{86.28}\vspace{0.1cm}\\
	  \bottomrule
	\end{tabular}
\tabcaption{\textbf{Part Segmentation on ShapeNetPart \protect\cite{shapenetpart}}. `mIoU$_C$' (\%) and `mIoU$_I$' (\%) denote the mean IoU across all part categories and all instances in the dataset, respectively.}
\label{seg}
\vspace{-0.1cm}
\end{figure}

\paragraph{Shape Classification.}
We utilize a simple a classification head of linear layers and evaluate the accuracy on ModelNet40~\cite{modelnet40} and ScanObjectNN~\cite{scanobjectnn} datasets, which contain synthetic objects and real-world instances, respectively. 
Table~\ref{modelnet_cls} demonstrates the classification results on ModelNet40, where we train on 9,843 instances and test on 2,468 instances with 40 categories. Joint-MAE achieves the highest accuracy compared with other methods. Note that we randomly sample 1,024 points from each object with only the coordinate information as the input data.
Besides, we report in Table~\ref{scan_cls} the performance on the real-world ScanObjectNN dataset, which consists of 2,304 objects for training and 576 objects for testing. Joint-MAE exceeds Point-MAE by an improvement of +0.89\% on the most challenging part, i.e., `PB-T50-RS' in Table~\ref{scan_cls}. As the pre-training dataset, ShapeNet, is composed of synthetic point clouds, this results indicate the superiority of Joint-MAE for out-of-distribution data.

\paragraph{Few-shot Classification.}
To evaluate the representation ability of our Joint-MAE on limited training data, we further conduct few-shot classification task on ModelNet40~\cite{modelnet40} as well. 
Following the common routine, we perform $N$-way $K$-shot experiments on Joint-MAE for 10 times, which means we randomly select $N$ classes from ModelNet40, and sample $K$ objects from each class. 
Table~\ref{fewshot} reports the comparison results, where we exhibit the mean and standard deviation over 10 runs.
As reported, our Joint-MAE outperforms prior works in almost all few-shot settings, indicating strong capacity under low-data regimes.

\paragraph{Part Segmentation.}
To evaluate the understanding ability of Joint-MAE on the fine-grained dense 3D data, we also conduct the part segmentation experiment on ShapeNetPart dataset~\cite{shapenetpart}, which contains 16,881 instances of 16 categories. Following Point-MAE~\cite{pointmae}, we sample 2,048 points from each input instance, and adopt the same segmentation head for fair comparison, which concatenates the output features from different transformer blocks of the encoder.
Table~\ref{seg} shows the comparison results of our Joint-MAE with existing supervised methods and self-supervised methods.
We here report the mean IoU across all part categories (denoted as $\text{mIoU}_C$) and all instances (denoted as $\text{mIoU}_I$), respectively.
Clearly, our Joint-MAE achieves the highest scores on both metrics. This well illustrates the superior dense 3D representation capacity of Joint-MAE learned from our 2D-3D cross-modal pre-training.

\subsection{Ablation Study}
To explore the contributions of each major component in Joint-MAE, we conduct extensive ablations on ModelNet40 and evaluate the performance by the Linear SVM classification accuracy.

\vspace{0.2cm}
\paragraph{2D-3D Embedding Modules and Joint Decoder.} \ 
On top of our final design, we conduct experiments by removing the hierarchical 2D-3D embedding modules and modal-shared decoder, respectively. As reported in Table~\ref{block}, the absence (denoted by `-') of either modal-specific embedding module hurts the performance. The results show the effectiveness of both our initial embedding modules for multi-scale learning and the modal-shared decoder for 2D-3D feature interaction for mask tokens.

\begin{table}[t!]
\vspace{0.15cm}
\centering
\small
	\begin{tabular}{ccccc}
	\toprule
 \multicolumn{2}{c}{\ \ \ \ \ \ \ \ \ \ \ \ \ \ \ \ Attention Schemes\ \ \ \ \ \ \ \ \ \ \ \ \ \ \ \ } &\makecell*[c]{\multirow{2}*{\ \ \ Acc. (\%)\ \ \ }} \\
 \cmidrule(lr){1-2} 
    \ \ \ \ \ \ \ 2D-3D\ \ \ \ \ \ \  &3D-2D & \\
		 \cmidrule(lr){1-1}  \cmidrule(lr){2-2} \cmidrule(lr){3-3}
            Global &Global &91.7\\
            Local &Global  &91.8\\
            Global &Local &92.0\\
            Local &Local &\textbf{92.4}\\
	  \bottomrule
	\end{tabular}
\tabcaption{\textbf{Ablation Results on Local-aligned Attention}. We utilize `Global’ and `Local’ to denote the standard global attention and local-aligned attention mechanism respectively. `2D-3D’ and `3D-2D’ denote the cross-modal positions during attention calculation.}
\vspace{-0.25cm}
\label{attn_abla}
\end{table}

\begin{figure}[t!]
\vspace{-0.1cm}
\centering
    \includegraphics[width=0.95\linewidth]{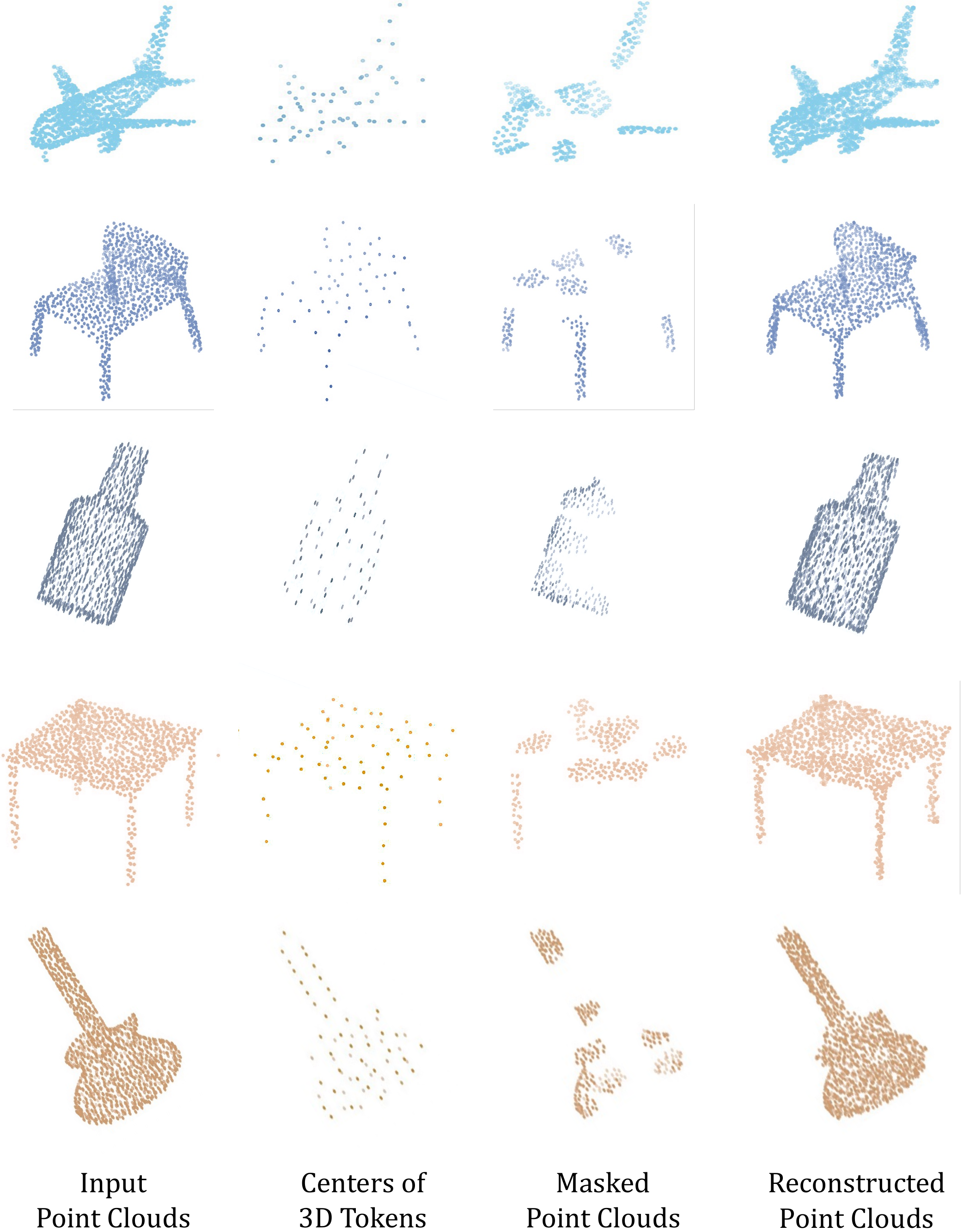}
    \vspace{0.1cm}
  \caption{\textbf{Visualization of Point Clouds from Joint-MAE.} In each row, we visualize the input point clouds,
the centers of 3D tokens, the masked point clouds, and the reconstructed coordinates.}
  \label{visual}
\end{figure}
\begin{table}[h!]
\vspace{-0.1cm}
\centering
\small
	\begin{tabular}{ccccc}
	\toprule
            \multicolumn{2}{c}{Embed. Modules} &\multicolumn{2}{c}{Joint Decoder} &\makecell*[c]{\multirow{2}*{Acc. (\%)}}\\
		 \cmidrule(lr){1-2}  \cmidrule(lr){3-4}
		 \ \ \ 2D\ \ \  &3D &Modal-shared &Modal-specific &\\
		 \cmidrule(lr){1-1} \cmidrule(lr){2-2} \cmidrule(lr){3-3} \cmidrule(lr){4-4} \cmidrule(lr){5-5}
            \checkmark &\checkmark &\checkmark &\checkmark & \textbf{92.4}\\
	    \checkmark &-  &\checkmark  &\checkmark &91.2\\
            - & \checkmark & \checkmark & \checkmark &91.4\\
            \checkmark &\checkmark &- &\checkmark &91.8\\
	  \bottomrule
	\end{tabular}
\tabcaption{\textbf{Ablation Results on Hierarchical Embedding Modules and Joint Decoder}. {`Acc.'} denotes the accuracy performance.}
\vspace{-0.17cm}
\label{block}
\end{table}

\paragraph{Local-aligned Attention.}
In Table~\ref{attn_abla}, we experiment different attention schemes for cross-modal positions shown in Figure~\ref{attn}, i.e., 2D-3D and 3D-2D correlative tokens. We denote the local-aligned attention and the original attention as `Local' and `Global', respectively. The last row represents our Joint-MAE with local-aligned attention for all cross-modal positions. Compared with others, the local-aligned attention mechanism contributes to +0.5\% classification accuracy, indicating that the proposed method can boost point cloud representation learning with 2D fine-grained semantic cues.

\begin{table}[t]
\centering
\small
	\begin{tabular}{cccc}
	\toprule
        \makecell*[c]{\multirow{2}*{\ \ \ \ Cross-Reconstruct.\ \ \ \ }} &\multicolumn{2}{c}{\ \ \ Proj. Settings\ \ \ } &\makecell*[c]{\multirow{2}*{\ \ Acc. (\%)\ \ }} \\
        \cmidrule(lr){2-3}
        &View Num. &Render & \\
		 \cmidrule(lr){1-1}  \cmidrule(lr){2-2} \cmidrule(lr){3-3} \cmidrule(lr){4-4}
            - &- &- & 91.1\\
            \checkmark &1 &- &91.6\\
            \checkmark &4 &- &\textbf{92.4}\\
            \checkmark &4 &\checkmark &91.9\\
	  \bottomrule
	\end{tabular}
\tabcaption{\textbf{Ablation Results on Loss Functions.} `View Num' denotes the view number of projections in the cross-reconstruction loss. `Render' denotes whether we conduct color rendering on the projected 2D depth maps.}
\vspace{-0.4cm}
\label{loss_abla}
\end{table}

\paragraph{Cross-reconstruction Loss.}
In Table~\ref{loss_abla}, we explore the projection settings of the cross-reconstruction loss. As reported, the proposed cross-reconstruction loss contributes to at least +0.5\% on the classification scores. The comparison of the middle two rows reveals that projecting from multiple perspectives, which enforces more geometric constraints, can well benefit 3D point cloud pre-training. Also, the performance decay in the last row indicates the rendering operation might introduce extra noises that disturb the effective supervision on the reconstructed 3D point clouds.

\section{Visualization}
In Figure~\ref{visual}, to intuitively verify the constructed performance of Joint-MAE, we visualize the input point clouds, the centers of 3D tokens ($N/32$ point number), the masked point clouds, and the reconstructed coordinates, respectively in each row. As shown, with the proposed 2D-3D joint framework and cross-modal learning strategies, the network can well generate masked point clouds with fine-grained local structures.


\section{Conclusion}
\label{sec:conclusion}
We propose Joint-MAE, a multi-modal masked autoencoding framework for 3D point cloud pre-training. With the hierarchical embedding modules and the joint 2D-3D transformer, Joint-MAE can well fuse the informative characters between 2D and 3D modalities within a unified MAE architecture. By equipping the local-aligned attention and cross-reconstruction loss, Joint-MAE further learns strong cross-modal knowledge for 3D point cloud learning. Extensive experiments are conducted to demonstrate the superiority of Joint-MAE on 3D point cloud representation capability. We expect Joint-MAE can inspire more works to further combine multi-modality learning with 3D point cloud MAE using a unified framework. As for limitations, although Joint-MAE has verified the 2D modality can effectively benefit 3D point cloud pre-training, we will further explore how 3D modality improves 2D MAE as the future work. We do not foresee a negative social impact from the proposed work.

\section*{Acknowledgements}
This work was supported by the National Key R\&D Program of China (Grant No. 2022YFE0200700), the Hong Kong Innovation and Technology Commission (Project No. MHP/086/21), the China National Natural Science Foundation (Project No. 62202182), and the Research Grants Council of the Hong Kong Special Administrative Region, China (Project Reference Number: T45-401/22-N).

\section*{Contribution Statement}
Ziyu Guo and Renrui Zhang equally contribute to this work. Xianzhi Li is the corresponding author.
{
\bibliographystyle{named}
\bibliography{jointmae}

\begin{thebibliography}{}

\bibitem[\protect\citeauthoryear{Afham \bgroup \em et al.\egroup }{2022}]{afham2022crosspoint}
Mohamed Afham, Isuru Dissanayake, Dinithi Dissanayake, Amaya Dharmasiri, Kanchana Thilakarathna, and Ranga Rodrigo.
\newblock Crosspoint: Self-supervised cross-modal contrastive learning for 3d point cloud understanding.
\newblock {\em arXiv preprint arXiv:2203.00680}, 2022.

\bibitem[\protect\citeauthoryear{Chang \bgroup \em et al.\egroup }{2015}]{chang2015shapenet}
Angel~X Chang, Thomas Funkhouser, Leonidas Guibas, Pat Hanrahan, Qixing Huang, Zimo Li, Silvio Savarese, Manolis Savva, Shuran Song, Hao Su, et~al.
\newblock Shapenet: An information-rich 3d model repository.
\newblock {\em arXiv preprint arXiv:1512.03012}, 2015.

\bibitem[\protect\citeauthoryear{Desai and Johnson}{2021}]{desai2021virtex}
Karan Desai and Justin Johnson.
\newblock Virtex: Learning visual representations from textual annotations.
\newblock In {\em Proceedings of the IEEE/CVF conference on computer vision and pattern recognition}, pages 11162--11173, 2021.

\bibitem[\protect\citeauthoryear{Fan \bgroup \em et al.\egroup }{2017}]{fan2017point}
Haoqiang Fan, Hao Su, and Leonidas~J Guibas.
\newblock A point set generation network for 3d object reconstruction from a single image.
\newblock In {\em Proceedings of the IEEE conference on computer vision and pattern recognition}, pages 605--613, 2017.

\bibitem[\protect\citeauthoryear{Gao \bgroup \em et al.\egroup }{2021}]{gao2021clip}
Peng Gao, Shijie Geng, Renrui Zhang, Teli Ma, Rongyao Fang, Yongfeng Zhang, Hongsheng Li, and Yu~Qiao.
\newblock Clip-adapter: Better vision-language models with feature adapters.
\newblock {\em arXiv preprint arXiv:2110.04544}, 2021.

\bibitem[\protect\citeauthoryear{Gao \bgroup \em et al.\egroup }{2022}]{gao2022convmae}
Peng Gao, Teli Ma, Hongsheng Li, Jifeng Dai, and Yu~Qiao.
\newblock Convmae: Masked convolution meets masked autoencoders.
\newblock {\em arXiv preprint arXiv:2205.03892}, 2022.

\bibitem[\protect\citeauthoryear{Gao \bgroup \em et al.\egroup }{2023}]{gaomimic}
Peng Gao, Renrui Zhang, Rongyao Fang, Ziyi Lin, Hongyang Li, Hongsheng Li, and Qiao Yu.
\newblock Mimic before reconstruct: Enhancing masked autoencoders with feature mimicking.
\newblock {\em arXiv preprint arXiv:2303.05475}, 2023.

\bibitem[\protect\citeauthoryear{Goyal \bgroup \em et al.\egroup }{2021}]{simpleview}
Ankit Goyal, Hei Law, Bowei Liu, Alejandro Newell, and Jia Deng.
\newblock Revisiting point cloud shape classification with a simple and effective baseline.
\newblock {\em arXiv preprint arXiv:2106.05304}, 2021.

\bibitem[\protect\citeauthoryear{Guo \bgroup \em et al.\egroup }{2021}]{guo2021pct}
Meng-Hao Guo, Jun-Xiong Cai, Zheng-Ning Liu, Tai-Jiang Mu, Ralph~R Martin, and Shi-Min Hu.
\newblock Pct: Point cloud transformer.
\newblock {\em Computational Visual Media}, 7(2):187--199, 2021.

\bibitem[\protect\citeauthoryear{Guo \bgroup \em et al.\egroup }{2022}]{guo2022calip}
Ziyu Guo, Renrui Zhang, Longtian Qiu, Xianzheng Ma, Xupeng Miao, Xuming He, and Bin Cui.
\newblock Calip: Zero-shot enhancement of clip with parameter-free attention.
\newblock {\em arXiv preprint arXiv:2209.14169}, 2022.

\bibitem[\protect\citeauthoryear{Han \bgroup \em et al.\egroup }{2019}]{vipgan}
Zhizhong Han, Mingyang Shang, Yu-Shen Liu, and Matthias Zwicker.
\newblock View inter-prediction gan: Unsupervised representation learning for 3d shapes by learning global shape memories to support local view predictions.
\newblock In {\em Proceedings of the AAAI Conference on Artificial Intelligence}, volume~33, pages 8376--8384, 2019.

\bibitem[\protect\citeauthoryear{He \bgroup \em et al.\egroup }{2021}]{mae}
Kaiming He, Xinlei Chen, Saining Xie, Yanghao Li, Piotr Doll{\'a}r, and Ross Girshick.
\newblock Masked autoencoders are scalable vision learners.
\newblock {\em arXiv preprint arXiv:2111.06377}, 2021.

\bibitem[\protect\citeauthoryear{Li \bgroup \em et al.\egroup }{2018}]{sonet}
Jiaxin Li, Ben~M Chen, and Gim~Hee Lee.
\newblock So-net: Self-organizing network for point cloud analysis.
\newblock In {\em Proceedings of the IEEE conference on computer vision and pattern recognition}, pages 9397--9406, 2018.

\bibitem[\protect\citeauthoryear{Liu \bgroup \em et al.\egroup }{2022}]{maskpoint}
Haotian Liu, Mu~Cai, and Yong~Jae Lee.
\newblock Masked discrimination for self-supervised learning on point clouds.
\newblock {\em arXiv preprint arXiv:2203.11183}, 2022.

\bibitem[\protect\citeauthoryear{Nichol \bgroup \em et al.\egroup }{2022}]{nichol2022point}
Alex Nichol, Heewoo Jun, Prafulla Dhariwal, Pamela Mishkin, and Mark Chen.
\newblock Point-e: A system for generating 3d point clouds from complex prompts.
\newblock {\em arXiv preprint arXiv:2212.08751}, 2022.

\bibitem[\protect\citeauthoryear{Pang \bgroup \em et al.\egroup }{2022}]{pointmae}
Yatian Pang, Wenxiao Wang, Francis~EH Tay, Wei Liu, Yonghong Tian, and Li~Yuan.
\newblock Masked autoencoders for point cloud self-supervised learning.
\newblock {\em arXiv preprint arXiv:2203.06604}, 2022.

\bibitem[\protect\citeauthoryear{Poursaeed \bgroup \em et al.\egroup }{2020}]{poursaeed2020self}
Omid Poursaeed, Tianxing Jiang, Han Qiao, Nayun Xu, and Vladimir~G Kim.
\newblock Self-supervised learning of point clouds via orientation estimation.
\newblock In {\em 2020 International Conference on 3D Vision (3DV)}, pages 1018--1028. IEEE, 2020.

\bibitem[\protect\citeauthoryear{Qi \bgroup \em et al.\egroup }{2017a}]{qi2017pointnet}
Charles~R Qi, Hao Su, Kaichun Mo, and Leonidas~J Guibas.
\newblock Pointnet: Deep learning on point sets for 3d classification and segmentation.
\newblock In {\em Proceedings of the IEEE conference on computer vision and pattern recognition}, pages 652--660, 2017.

\bibitem[\protect\citeauthoryear{Qi \bgroup \em et al.\egroup }{2017b}]{qi2017pointnet++}
Charles~R Qi, Li~Yi, Hao Su, and Leonidas~J Guibas.
\newblock Pointnet++: Deep hierarchical feature learning on point sets in a metric space.
\newblock {\em arXiv preprint arXiv:1706.02413}, 2017.

\bibitem[\protect\citeauthoryear{Radford \bgroup \em et al.\egroup }{2021}]{clip}
Alec Radford, Jong~Wook Kim, Chris Hallacy, Aditya Ramesh, Gabriel Goh, Sandhini Agarwal, Girish Sastry, Amanda Askell, Pamela Mishkin, Jack Clark, et~al.
\newblock Learning transferable visual models from natural language supervision.
\newblock In {\em International Conference on Machine Learning}, pages 8748--8763. PMLR, 2021.

\bibitem[\protect\citeauthoryear{Rao \bgroup \em et al.\egroup }{2020}]{rao2020global}
Yongming Rao, Jiwen Lu, and Jie Zhou.
\newblock Global-local bidirectional reasoning for unsupervised representation learning of 3d point clouds.
\newblock In {\em Proceedings of the IEEE/CVF Conference on Computer Vision and Pattern Recognition}, pages 5376--5385, 2020.

\bibitem[\protect\citeauthoryear{Sauder and Sievers}{2019}]{jiasaw}
Jonathan Sauder and Bjarne Sievers.
\newblock Self-supervised deep learning on point clouds by reconstructing space.
\newblock {\em Advances in Neural Information Processing Systems}, 32, 2019.

\bibitem[\protect\citeauthoryear{Uy \bgroup \em et al.\egroup }{2019}]{scanobjectnn}
Mikaela~Angelina Uy, Quang-Hieu Pham, Binh-Son Hua, Thanh Nguyen, and Sai-Kit Yeung.
\newblock Revisiting point cloud classification: A new benchmark dataset and classification model on real-world data.
\newblock In {\em Proceedings of the IEEE/CVF International Conference on Computer Vision}, pages 1588--1597, 2019.

\bibitem[\protect\citeauthoryear{Vaswani \bgroup \em et al.\egroup }{2017}]{transformer}
Ashish Vaswani, Noam Shazeer, Niki Parmar, Jakob Uszkoreit, Llion Jones, Aidan~N Gomez, {\L}ukasz Kaiser, and Illia Polosukhin.
\newblock Attention is all you need.
\newblock In {\em Advances in neural information processing systems}, pages 5998--6008, 2017.

\bibitem[\protect\citeauthoryear{Wang \bgroup \em et al.\egroup }{2019}]{dgcnn}
Yue Wang, Yongbin Sun, Ziwei Liu, Sanjay~E Sarma, Michael~M Bronstein, and Justin~M Solomon.
\newblock Dynamic graph cnn for learning on point clouds.
\newblock {\em Acm Transactions On Graphics (tog)}, 38(5):1--12, 2019.

\bibitem[\protect\citeauthoryear{Wang \bgroup \em et al.\egroup }{2021}]{occo}
Hanchen Wang, Qi~Liu, Xiangyu Yue, Joan Lasenby, and Matt~J Kusner.
\newblock Unsupervised point cloud pre-training via occlusion completion.
\newblock In {\em Proceedings of the IEEE/CVF International Conference on Computer Vision}, pages 9782--9792, 2021.

\bibitem[\protect\citeauthoryear{Wu \bgroup \em et al.\egroup }{2015a}]{modelnet40}
Zhirong Wu, Shuran Song, Aditya Khosla, Fisher Yu, Linguang Zhang, Xiaoou Tang, and Jianxiong Xiao.
\newblock 3d shapenets: A deep representation for volumetric shapes.
\newblock In {\em Proceedings of the IEEE conference on computer vision and pattern recognition}, pages 1912--1920, 2015.

\bibitem[\protect\citeauthoryear{Wu \bgroup \em et al.\egroup }{2015b}]{wu20153d}
Zhirong Wu, Shuran Song, Aditya Khosla, Fisher Yu, Linguang Zhang, Xiaoou Tang, and Jianxiong Xiao.
\newblock 3d shapenets: A deep representation for volumetric shapes.
\newblock In {\em Proceedings of the IEEE conference on computer vision and pattern recognition}, pages 1912--1920, 2015.

\bibitem[\protect\citeauthoryear{Xie \bgroup \em et al.\egroup }{2016}]{xie2016theory}
Jianwen Xie, Yang Lu, Song-Chun Zhu, and Yingnian Wu.
\newblock A theory of generative convnet.
\newblock In {\em International Conference on Machine Learning}, pages 2635--2644. PMLR, 2016.

\bibitem[\protect\citeauthoryear{Xie \bgroup \em et al.\egroup }{2018}]{xie2018learning}
Jianwen Xie, Zilong Zheng, Ruiqi Gao, Wenguan Wang, Song-Chun Zhu, and Ying~Nian Wu.
\newblock Learning descriptor networks for 3d shape synthesis and analysis.
\newblock In {\em Proceedings of the IEEE conference on computer vision and pattern recognition}, pages 8629--8638, 2018.

\bibitem[\protect\citeauthoryear{Xie \bgroup \em et al.\egroup }{2020}]{xie2020generative}
Jianwen Xie, Zilong Zheng, Ruiqi Gao, Wenguan Wang, Song-Chun Zhu, and Ying~Nian Wu.
\newblock Generative voxelnet: learning energy-based models for 3d shape synthesis and analysis.
\newblock {\em IEEE Transactions on Pattern Analysis and Machine Intelligence}, 44(5):2468--2484, 2020.

\bibitem[\protect\citeauthoryear{Xie \bgroup \em et al.\egroup }{2021}]{xie2021generative}
Jianwen Xie, Yifei Xu, Zilong Zheng, Song-Chun Zhu, and Ying~Nian Wu.
\newblock Generative pointnet: Deep energy-based learning on unordered point sets for 3d generation, reconstruction and classification.
\newblock In {\em Proceedings of the IEEE/CVF Conference on Computer Vision and Pattern Recognition}, pages 14976--14985, 2021.

\bibitem[\protect\citeauthoryear{Xu \bgroup \em et al.\egroup }{2021}]{xu2021image2point}
Chenfeng Xu, Shijia Yang, Bohan Zhai, Bichen Wu, Xiangyu Yue, Wei Zhan, Peter Vajda, Kurt Keutzer, and Masayoshi Tomizuka.
\newblock Image2point: 3d point-cloud understanding with pretrained 2d convnets.
\newblock {\em arXiv preprint arXiv:2106.04180}, 2021.

\bibitem[\protect\citeauthoryear{Yang \bgroup \em et al.\egroup }{2018}]{foldingnet}
Yaoqing Yang, Chen Feng, Yiru Shen, and Dong Tian.
\newblock Foldingnet: Point cloud auto-encoder via deep grid deformation.
\newblock In {\em Proceedings of the IEEE conference on computer vision and pattern recognition}, pages 206--215, 2018.

\bibitem[\protect\citeauthoryear{Yi \bgroup \em et al.\egroup }{2016}]{shapenetpart}
Li~Yi, Vladimir~G Kim, Duygu Ceylan, I-Chao Shen, Mengyan Yan, Hao Su, Cewu Lu, Qixing Huang, Alla Sheffer, and Leonidas Guibas.
\newblock A scalable active framework for region annotation in 3d shape collections.
\newblock {\em ACM Transactions on Graphics (ToG)}, 35(6):1--12, 2016.

\bibitem[\protect\citeauthoryear{Yu \bgroup \em et al.\egroup }{2021}]{pointbert}
Xumin Yu, Lulu Tang, Yongming Rao, Tiejun Huang, Jie Zhou, and Jiwen Lu.
\newblock Point-bert: Pre-training 3d point cloud transformers with masked point modeling.
\newblock {\em arXiv preprint arXiv:2111.14819}, 2021.

\bibitem[\protect\citeauthoryear{Zhang \bgroup \em et al.\egroup }{2021a}]{tipadapter}
Renrui Zhang, Rongyao Fang, Wei Zhang, Peng Gao, Kunchang Li, Jifeng Dai, Yu~Qiao, and Hongsheng Li.
\newblock Tip-adapter: Training-free clip-adapter for better vision-language modeling.
\newblock {\em arXiv preprint arXiv:2111.03930}, 2021.

\bibitem[\protect\citeauthoryear{Zhang \bgroup \em et al.\egroup }{2021b}]{zhang2021pointclip}
Renrui Zhang, Ziyu Guo, Wei Zhang, Kunchang Li, Xupeng Miao, Bin Cui, Yu~Qiao, Peng Gao, and Hongsheng Li.
\newblock Pointclip: Point cloud understanding by clip.
\newblock {\em arXiv preprint arXiv:2112.02413}, 2021.

\bibitem[\protect\citeauthoryear{Zhang \bgroup \em et al.\egroup }{2021c}]{zhang2021self}
Zaiwei Zhang, Rohit Girdhar, Armand Joulin, and Ishan Misra.
\newblock Self-supervised pretraining of 3d features on any point-cloud.
\newblock In {\em Proceedings of the IEEE/CVF International Conference on Computer Vision}, pages 10252--10263, 2021.

\bibitem[\protect\citeauthoryear{Zhang \bgroup \em et al.\egroup }{2022a}]{pointm2ae}
Renrui Zhang, Ziyu Guo, Peng Gao, Rongyao Fang, Bin Zhao, Dong Wang, Yu~Qiao, and Hongsheng Li.
\newblock Point-m2ae: Multi-scale masked autoencoders for hierarchical point cloud pre-training.
\newblock {\em arXiv preprint arXiv:2205.14401}, 2022.

\bibitem[\protect\citeauthoryear{Zhang \bgroup \em et al.\egroup }{2022b}]{i2p}
Renrui Zhang, Liuhui Wang, Yu~Qiao, Peng Gao, and Hongsheng Li.
\newblock Learning 3d representations from 2d pre-trained models via image-to-point masked autoencoders.
\newblock {\em arXiv preprint arXiv:2212.06785}, 2022.

\bibitem[\protect\citeauthoryear{Zhang \bgroup \em et al.\egroup }{2022c}]{depthclip}
Renrui Zhang, Ziyao Zeng, Ziyu Guo, and Yafeng Li.
\newblock Can language understand depth?
\newblock In {\em Proceedings of the 30th ACM International Conference on Multimedia}, pages 6868--6874, 2022.

\bibitem[\protect\citeauthoryear{Zhang \bgroup \em et al.\egroup }{2023}]{zhang2023nearest}
Renrui Zhang, Liuhui Wang, Ziyu Guo, and Jianbo Shi.
\newblock Nearest neighbors meet deep neural networks for point cloud analysis.
\newblock In {\em Proceedings of the IEEE/CVF Winter Conference on Applications of Computer Vision}, pages 1246--1255, 2023.

\bibitem[\protect\citeauthoryear{Zhao \bgroup \em et al.\egroup }{2021}]{pointtransformer}
Hengshuang Zhao, Li~Jiang, Jiaya Jia, Philip~HS Torr, and Vladlen Koltun.
\newblock Point transformer.
\newblock In {\em Proceedings of the IEEE/CVF International Conference on Computer Vision}, pages 16259--16268, 2021.

\bibitem[\protect\citeauthoryear{Zhu \bgroup \em et al.\egroup }{2022}]{pointclipv2}
Xiangyang Zhu, Renrui Zhang, Bowei He, Ziyao Zeng, Shanghang Zhang, and Peng Gao.
\newblock Pointclip v2: Adapting clip for powerful 3d open-world learning.
\newblock {\em arXiv preprint arXiv:2211.11682}, 2022.

\end{thebibliography}
}

\end{document}


\maketitle

\section{Implementation Details}
\paragraph{Positional Encodings.}
\label{position}
To complement the positional information of two modalities in the transformers, we adopt learnable positional encodings to all attention layers in Joint-MAE. 
For 2D modality, we initialize the encodings with sine-cosine function and keep updating them during pre-training. For 3D modality, we utilize a two-layer MLP to encode the 3D coordinates of point clouds into $C$-dimension vectors. Then, we respectively add the positional encodings of two modalities into the features by element before feeding into the joint encoder.

\begin{table}[t!]
\centering
\small
\tabcaption{\textbf{Ablations on Block Number}. {`Acc.'} denotes the classification accuracy.}
\vspace{-0.1cm}
\label{block}
	\begin{tabular}{cccc}
	\toprule
            \makecell*[c]{\multirow{2}*{Joint Encoder}} &\multicolumn{2}{c}{Joint Decoder} &\makecell*[c]{\multirow{2}*{Acc. (\%)}}\\
		\cmidrule(lr){2-3}
		  &Modal-shared &Modal-specific &\\
		 \cmidrule(lr){1-1} \cmidrule(lr){2-2} \cmidrule(lr){3-3} \cmidrule(lr){4-4} 
            8 &2 &1 &91.9 \\
            10 &2 &1 &92.0 \\
            12 &1 &1 &92.3 \\
            12 &2 &2 &91.7 \\
            12 &2 &1 &\textbf{92.4} \\
	  \bottomrule
	\end{tabular}
\vspace{-0.1cm}
\end{table}

\paragraph{Self-supervised Pre-training.}
As presented in Section \textcolor{linkcolor}{4.1}, we pre-train our Joint-MAE on ShapeNet~\cite{chang2015shapenet} dataset.
We pre-train the network for 400 epochs with a batch size 128. We utilize AdamW~\cite{kingma2014adam} as the optimizer with an initial learning rate of 5$\times$10$^{-5}$ and a weight decay as 5$\times$10$^{-2}$. We adopt the cosine scheduler along with a 10-epoch warm-up. To project 2D depth maps, we scale the point clouds into $[1, −1]^3$, and keep the cameras at a distance of 1.4 units.
As for data augmentation, we utilize random scaling, rotating, translation and jitter for 3D point clouds during pre-training. We do not adopt extra data augmentation on projected 2D depth maps, since projection from random view servers as an augmentation of rotating. For the hierarchical 3D embedding, we set the point numbers ($N/8$, $N/32$) and token dimensions as \{128, 32\} and \{192, 384\}, and adopt k for the k-NN as \{16, 4\}. 
We utilize a patch size of 16 for 2D pathifying, and the hierarchical 2D embedding consists of two convolution blocks with output spatial resolutions of \{$H/4 \times W/4$, $H/16 \times W/16$\} and dimensions of \{128, 384\}, respectively, where $H \times W$ is the resolution of 2D input. Considering the multi-scale circumstance, we utilize multi-scale masking~\cite{pointm2ae} and block-wise masking~\cite{gao2022convmae} respectively for 2D and 3D modalities, whose ratio is 75\%. We adopt 12 transformer blocks in the joint encoder, followed by a 2-block modal-shared decoder, and 1-block 2D/3D specific decoder, and we set 6 heads for the attention modules in both the joint encoder and the joint decoder.

\noindent\begin{figure}[t!]
    \centering
    \includegraphics[width=0.85\linewidth]{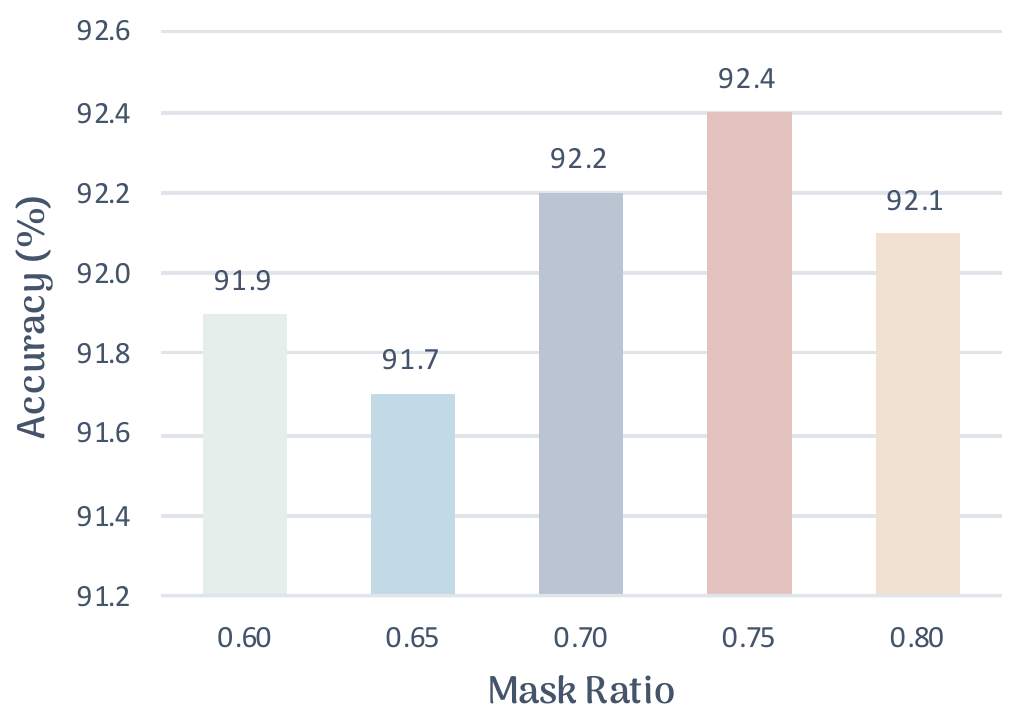}
  \caption{\textbf{Ablations on Mask Ratio}. The mask ratio 75\% performs the best for Joint-MAE.}
  \label{ratio}
\end{figure}

\begin{figure*}[t!]
\vspace{-0.35cm}
  \centering
    \includegraphics[width=0.9\textwidth]{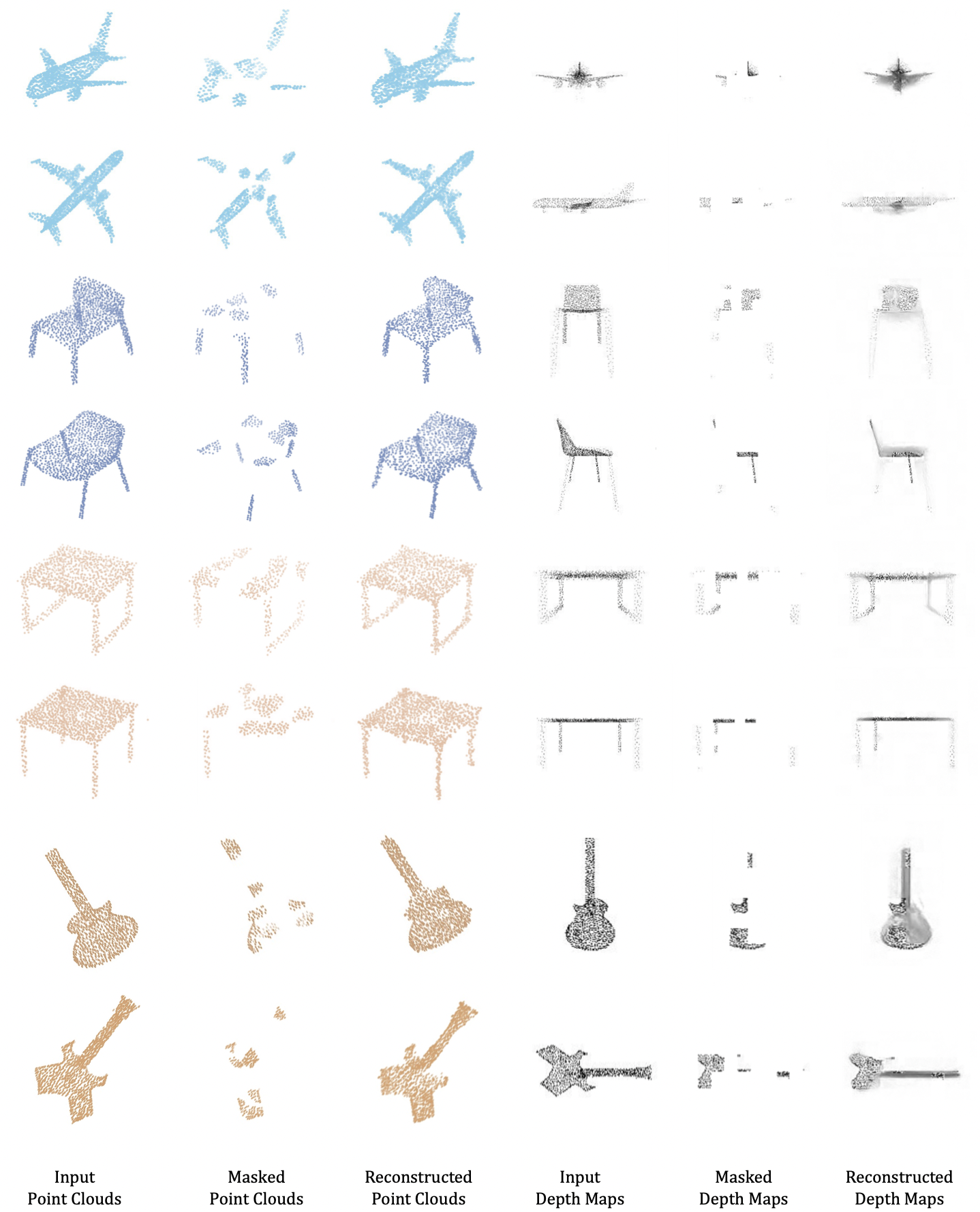}
   \caption{\textbf{Visualization of Point Clouds and Depth Maps from Joint-MAE.} In each column, we visualize the input point clouds, the masked point clouds, the reconstructed coordinates, the input projected depth maps, the masked depth maps, and the reconstructed depth maps.}
    \label{sup_vis}
\end{figure*}

\paragraph{Linear SVM.}
For linear SVM on ModelNet40~\cite{modelnet40}, we utilize both max and average pooling operations to aggregate the 3D tokens from the pre-trained encoder, and sum the two results as the global feature for SVM classification.

\paragraph{Shape Classification.}
We fine-tune the 3D-branch encoder of Joint-MAE on two datasets for shape classification: the widely adopted ModelNet40~\cite{modelnet40} and the challenging ScanObjectNN~\cite{scanobjectnn}, with synthetic 3D shapes and objects from noisy real-world scenes, respectively.
Referring to previous works, we randomly sample 1,024 and 2,048 points for each object from the two datasets. For fair comparison, we follow existing works to utilize the voting strategy~\cite{rscnn} on ModelNet40, but not ScanObjectNN. 
For both datasets, we fine-tune the model for 300 epochs with a batch size 32, and set the learning rate as 1$\times$10$^{-4}$. We keep other hyper-parameters of Joint-MAE the same as the pre-training process.

\paragraph{Few-shot Classification.}
Following previous works~\cite{pointbert,afham2022crosspoint,occo}, we evaluate the few-shot classification performance of Joint-MAE under the ``\textit{K}-way \textit{N}-shot'' settings. We randomly select \textit{K} classes from 40 and sample $N$+20 objects per class, where $N$ for training and the other 20 3D shapes for testing. We only fine-tune the network for 150 epochs, and keep other hyper-parameters the same as in the shape classification experiments.


\paragraph{Part Segmentation}
We conduct part segmentation experiments on ShapeNetPart~\cite{shapenetpart} that contains 14,007 and 2,874 samples for training and validation, respectively. We fine-tune Joint-MAE for 300 epochs with a batch size of 16. We adopt the learning rate as 1$\times$10$^{-4}$ and the weight decay as 0.1. Other training settings are the same as in the shape classification experiments.
\vspace{-0.1cm}
\section{Additional Ablation Study}
We further conduct ablation studies on pre-training and downstream fine-tuning. For the first three ablations, we utilize our final version as the baseline and compare the linear SVM classification accuracy (\%) on ModelNet40~\cite{modelnet40}.


\begin{table}[t!]
\centering
\small
\tabcaption{\textbf{With and without the pre-training.} `ModelNet40 Cls.' and `ScanObjectNN Cls.' denote shape classification on ModelNet40~\cite{modelnet40} and ScanObjectNN~\cite{scanobjectnn}, respectively. `Few-shot Cls.' represents the few-shot classification on 10-way 20-shot ModelNet40, and `Part Seg.' denotes part segmentation on ShapeNetPart~\cite{shapenetpart}.}
\begin{tabular}{lccc}
    \toprule
    \ Downstream Task &w/o (\%) &w (\%) &Gain\\
    \cmidrule(lr){1-1} \cmidrule(lr){2-2} \cmidrule(lr){3-3} \cmidrule(lr){4-4}
     \ ModelNet40 Cls. &91.9 &94.0 &\textcolor{violet}{+2.1\%}\vspace{0.05cm} \\
     \ ScanObjectNN Cls. &84.2 &86.1 &\textcolor{violet}{+1.9\%}\vspace{0.05cm} \\
     \ Few-shot Cls. &91.3 &95.1 &\textcolor{violet}{+3.8\%}\vspace{0.05cm} \\
     \ Part Seg. &84.9 &86.3 &\textcolor{violet}{+1.4\%}\vspace{0.05cm} \\
    \bottomrule
\end{tabular}
\label{wwo}
\vspace{-0.2cm}
\end{table}

\paragraph{Transformer Blocks.}
We experiment with different block numbers in the joint encoder and the joint decoder in Table~\ref{block}. We observe that adopting a 12-block joint encoder, a 2-block modal-shared decoder, and 1-block 2D/3D specific decoders achieves the highest accuracy. With such asymmetric encoder-decoder architecture, the joint encoder can embed more semantic information from the two modalities, which benefits its representation ability and transfer capacity.

\paragraph{Mask Ratio.}
We pre-train Joint-MAE under different mask ratios, i.e., from 60$\%$ to $80\%$. As shown in Figure~\ref{ratio}, we observe that the best mask ratio is 75\%. Note that we utilize the same mask ratio on both modalities for the semantic alignment between 2D images and 3D point clouds.

\paragraph{Positional Encodings.}
\begin{wraptable}{r}{4.7cm}
\centering
\small
\vspace{-0.35cm}
\tabcaption{\textbf{Ablations on 2D Positional Encodings}.} 
\label{2dposition}
\begin{adjustbox}{width=\linewidth}
\begin{tabular}{lc}
\toprule
         {Type} &{Acc. (\%)} \\
        \cmidrule(lr){1-1} \cmidrule(lr){2-2}
        MLP &92.2\% \\
        Sin-Cos, Frozen &91.8\%\\
        Sin-Cos, Learnable &\textbf{92.4\%}\\
  \bottomrule
\end{tabular}
\end{adjustbox}
\end{wraptable}

We conduct ablation studies on 2D positional encodings. In Table~\ref{2dposition}, `Sin-Cos, Frozen' and `Sin-Cos, Learnable' represent the vectors initialized with the sin-cos function that are frozen and learnable, respectively. The `MLP' in the first row denotes utilizing a two-layer MLP to encode the 2D positions into $C$-dimensional vectors. Note that we fix the 3D positional encodings as vectors generated from a two-layer MLP, which is described in Section~\ref{position}. As reported, the `Sin-Cos, Frozen' performs the best.

\paragraph{With and Without Pre-training.}
In Table~\ref{wwo}, we report the performance of Join-MAE on downstream tasks with and without the pre-training. The `w/o' denotes that we randomly initialize the network and train it from scratch. As reported, the Joint-MAE pre-training boosts the performance on all downstream tasks by +2.1$\%$, +1.9$\%$, +3.8$\%$, and +1.4$\%$ respectively, which indicates the superiority of 2D-3D joint MAE pre-training.

\section{Additional Visualization}
\paragraph{Input and Reconstructed Data.} In each column of Figure~\ref{sup_vis}, we show the input point clouds, the masked point clouds, the reconstructed coordinates, the input projected depth maps, the masked depth maps, and the reconstructed depth maps, respectively. As shown, with the proposed 2D-3D joint MAE pre-training and cross-modal learning strategies, Joint-MAE can well generate masked point clouds and 2D depth maps.

\begin{figure}[t!]
    \centering
    \includegraphics[width=\linewidth]{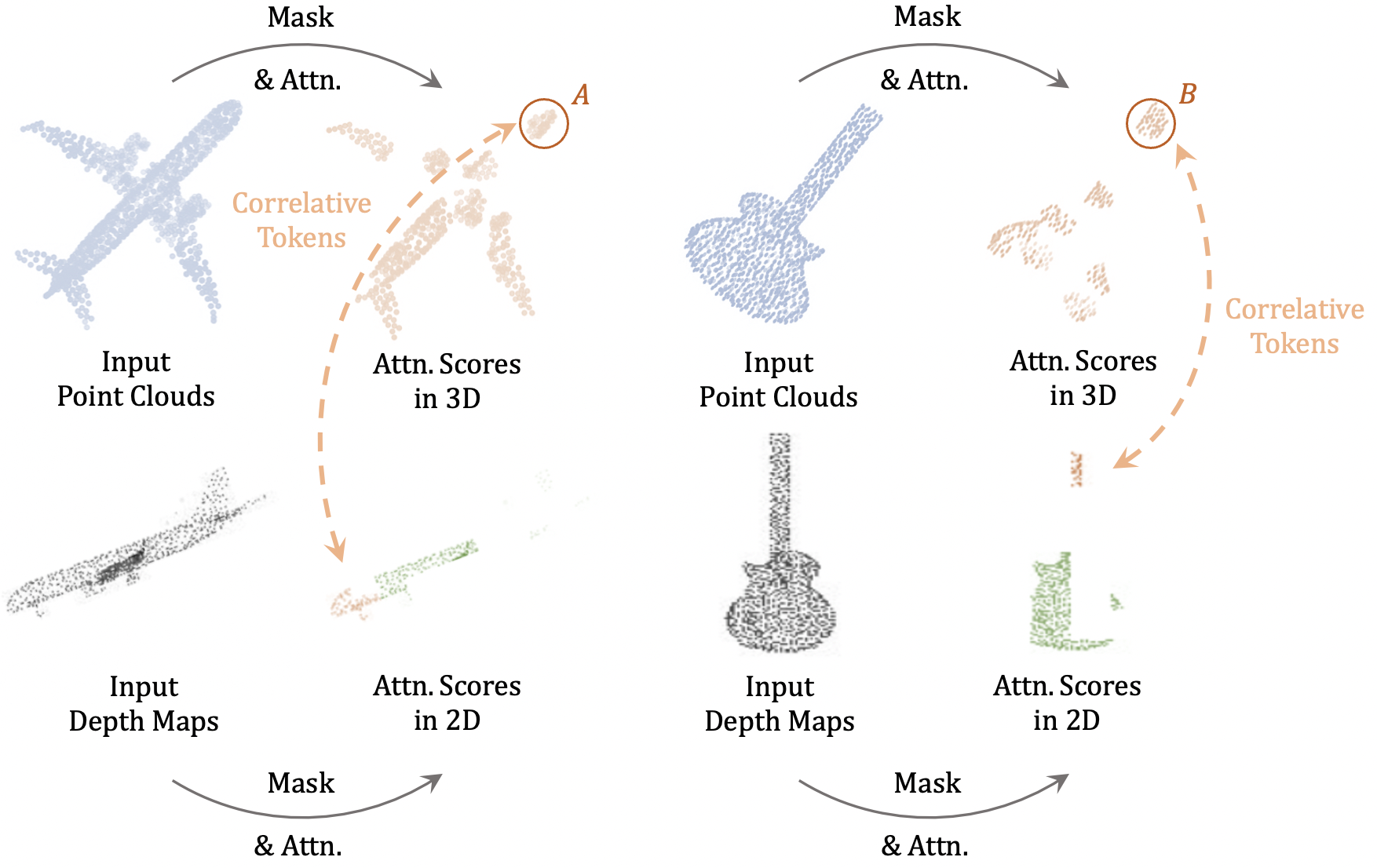}
    \vspace{0.07cm}
  \caption{\textbf{Visualization of Local-aligned Attention}. We visualize the attention scores with local-aligned attention. The query tokens are $A,B$ of 3D. The position in green means the attention weight is zero, while orange denotes nonzero, which are corresponding to the invalid-attention positions and valid-attention positions in Figure \textcolor{linkcolor}{3}.}
  \label{sup_attn}
  \vspace{0.2cm}
\end{figure}

\paragraph{Local-aligned Attention.}
We visualize the attention scores with local-aligned attention in Figure~\ref{sup_attn}. 
The input 3D point clouds and projected 2D depth maps are shown in the 1st and 3rd columns. 
In the 2nd and 4th columns of Figure~\ref{sup_attn}, the position in green means the attention weight here is zero, while orange denotes nonzero.
As shown, with the local-aligned attention, the query tokens $A,B$ of 3D focus on the whole 3D positions and only correlative parts of 2D depth maps, which are corresponding to the valid-attention positions in Section \textcolor{linkcolor}{3.3}. Meanwhile, the ignored positions in green are corresponding to the invalid-attention positions in Section \textcolor{linkcolor}{3.3}.
This local-aligned attention mechanism contributes to more fine-grained and intensive feature learning.

{
\bibliographystyle{named}
\bibliography{jointmae}
}